\documentclass[11pt,a4paper]{article}

\usepackage[utf8]{inputenc}
\usepackage[T1]{fontenc}
\usepackage{lmodern}
\usepackage{microtype}
\usepackage{amsmath,amssymb,amsthm}
\usepackage{graphicx}
\usepackage{booktabs}
\usepackage{multirow}
\usepackage{enumitem}
\usepackage{xcolor}
\usepackage{tikz}
\usetikzlibrary{arrows.meta,positioning}
\usepackage{xurl}
\usepackage[colorlinks=true,linkcolor=blue,citecolor=blue,urlcolor=blue]{hyperref}
\usepackage[numbers,sort&compress]{natbib}

\title{LLM StructCore:\\
Schema-Guided Reasoning Condensation and Deterministic Compilation}

\author{%
Serhii Zabolotnii\\
Cherkasy State Business College\\
Cherkasy, Ukraine\\
\texttt{zabolotnii.serhii@csbc.edu.ua}
}

\date{}

\begin{document}
\maketitle

\begin{abstract}
Automatically filling Case Report Forms (CRFs) from clinical notes is challenging due to noisy language, strict output contracts, and the high cost of false positives.
We describe our CL4Health 2026 submission for Dyspnea CRF filling (134~items) using a \textit{contract-driven} two-stage design grounded in \textit{Schema-Guided Reasoning} (SGR)~\cite{abdullin2026sgr}.
The key task property is \textbf{extreme sparsity}: the majority of fields are \texttt{unknown}, and official scoring penalizes both empty values and unsupported predictions.
We shift from a single-step ``LLM predicts 134~fields'' approach to a decomposition where
(i)~Stage~1 produces a stable SGR-style JSON summary with exactly 9~domain keys, and
(ii)~Stage~2 is a fully deterministic, 0-LLM compiler that parses the Stage~1 summary, canonicalizes item names (optionally using a UMLS alias map with 134/134 coverage), normalizes predictions to the official controlled vocabulary (13~categories), applies evidence-gated false-positive filters, and expands the output into the required 134-item format.
On the dev80 split, the best teacher configuration (Mistral Large~3 Stage~1 $\rightarrow$ Stage~2 deterministic) achieves macro-F1 \textbf{0.6543}~(EN) and \textbf{0.6905}~(IT);
on the hidden test200, the submitted English variant scores \textbf{0.63} on Codabench.
The pipeline is language-agnostic: Italian results match or exceed English with no language-specific engineering.
\end{abstract}

\noindent\textit{Context:} This work was carried out in the context of the CRF Filling shared-task submission accepted to the Third Workshop on Patient-oriented Language Processing (CL4Health), co-located with LREC-COLING 2026.


% ─────────────────────────────────────────────
\section{Introduction}
\label{sec:intro}

The CL4Health 2026 CRF filling shared task requires mapping unstructured clinical notes to a strict 134-item Dyspnea CRF\@.
In practice, end-to-end LLM extraction often fails either due to output format drift (invalid JSON / JSONL structure) or due to unsupported ``hallucinated'' fills that increase false positives~(FP).
Our design goal is to separate \textbf{(a)}~clinical information condensation from \textbf{(b)}~contract enforcement and controlled-vocabulary normalization, while using SGR to make the intermediate representation stable and machine-checkable.

\paragraph{Schema-Guided Reasoning (SGR).}
SGR~\cite{abdullin2026sgr} is an architectural pattern that constrains an LLM to produce typed, schema-conformant outputs via structured output or constrained decoding~\cite{dong2024xgrammar,willard2023}, turning domain knowledge into executable data contracts.
This reduces variability and improves reliability compared to free-form text prompting.
In our system, Stage~1 uses an SGR-style schema with three core patterns to produce a stable 9-key domain summary:

\begin{itemize}[nosep]
    \item \textbf{Cascade (Domain Scaffold):} The LLM follows a fixed sequence of 9~clinical domains (Demographics $\rightarrow$ Vitals $\rightarrow$ Labs $\rightarrow$ Problems $\rightarrow$ Symptoms $\rightarrow$ Medications $\rightarrow$ Procedures $\rightarrow$ Utilization $\rightarrow$ Disposition) to ensure uniform coverage of the clinical narrative.
    \item \textbf{Cycle (Checklists):} The prompt enforces implicit checklists within each domain (e.g., examining 28~specific CRF problems, all critical vitals) to improve recall on sparse items.
    \item \textbf{Routing (Vocabulary):} While Stage~1 emits unconstrained text strings within the JSON structure, Stage~2 deterministically routes each prediction into one of 13~required controlled-vocabulary categories (Table~\ref{tab:vocab}).
\end{itemize}

Stage~2 then deterministically compiles this summary into the official CRF submission format, ensuring 100\% structural compliance.

% ─────────────────────────────────────────────
\section{Task and Data}
\label{sec:data}

\subsection{Datasets and Splits}
We use the official Hugging Face datasets~\cite{ferrazzi2025converting}
(\nolinkurl{NLP-FBK/dyspnea-crf-*}) and, for teacher generation, the in-domain
unannotated clinical notes set (\nolinkurl{NLP-FBK/dyspnea-clinical-notes},
2{,}667~records from SGB hospital, Italy).
Train contains 10~annotated records, development 80, and test 200 (hidden ground truth).

A key discovery is \textbf{extreme sparsity}: across the development set, the median number of non-\texttt{unknown} items per document is only $\sim$12 out of 134~total.
The most frequent non-unknown items are in VITALS (\textit{heart rate}, \textit{blood pressure}, \textit{spo2}) and LABS (\textit{hemoglobin}, \textit{creatinine}), while many diagnosis and procedure items appear in $<$5\% of documents.
This sparsity means that \texttt{unknown} is by far the most common class, and correctly predicting \texttt{unknown} for absent items is as important as correctly extracting present values.
The official scorer~\cite{ferrazzi2026} uses macro-F1 over items, which heavily penalizes both false positives (FP: predicting a value for an \texttt{unknown} item) and false negatives (FN: predicting \texttt{unknown} for a present item).

\subsection{Contract-Driven Dataset Parsing}
We align our pipeline contracts directly to the dataset schema:
(i)~the ordered list of 134~items is inferred from \texttt{annotations[*].item} and reused end-to-end;
(ii)~document identifiers are normalized to a plain ID (prefix before the first underscore) for internal alignment, and suffixed with the language tag (\texttt{\_en}/\texttt{\_it}) only for final submissions;
(iii)~intermediate representations are sparse, but the final builder always expands to all 134~items, filling missing values as \texttt{unknown}.

% ─────────────────────────────────────────────
\section{System Overview}
\label{sec:system}

Our pipeline follows a strict two-stage decomposition:

\smallskip
\begin{center}
\small
\texttt{Clinical Note}~$\rightarrow$~\texttt{Stage\,1 (LLM, SGR JSON)}\\
$\rightarrow$~\texttt{Stage\,2 (0-LLM compiler)}~$\rightarrow$~\texttt{Submission.jsonl}
\end{center}
\smallskip

\begin{figure}[t]
\centering
\begin{tikzpicture}[
    node distance=0.7cm and 0.2cm,
    box/.style={draw, rounded corners=3pt, align=center, fill=gray!8,
                text width=3.0cm, minimum height=0.9cm, font=\small},
    arrow/.style={-{Stealth[length=5pt]}, thick, gray!70}
]
\node[box, fill=white, draw=black!60] (note)
    {Clinical\\Note (\texttt{en}/\texttt{it})};
\node[box, below=of note, fill=blue!8, draw=blue!50] (stage1)
    {Stage 1 (LLM)\\SGR JSON\\(9 keys)};
\node[box, below=of stage1, fill=green!8, draw=green!50!black] (stage2)
    {Stage 2 (0-LLM)\\Deterministic\\Compiler};
\node[box, below=of stage2, fill=orange!8, draw=orange!60] (sub)
    {Submission\\(134 items)};

\draw[arrow] (note) -- (stage1);
\draw[arrow] (stage1) -- (stage2);
\draw[arrow] (stage2) -- (sub);

% Annotations
\node[anchor=west, text width=3.9cm, font=\scriptsize, align=left]
  at ([xshift=0.3cm, yshift=0.0cm]stage1.east)
  {\textbf{SGR Contract}\\
   -- Cascade: 9 domain sequence\\
   -- Cycle: checklists per domain\\
   -- Evidence-only facts\\
   -- JSON parse: 100\% stable};
\node[anchor=west, text width=3.9cm, font=\scriptsize, align=left]
  at ([xshift=0.3cm, yshift=0.0cm]stage2.east)
  {\textbf{4-Step Algorithm}\\
   1. KV extraction\\
   2. Canonicalization + UMLS\\
   3. FP-gates \& derivation\\
   4. Vocab norm (13 cat.)};

\end{tikzpicture}
\caption{The LLM StructCore two-stage pipeline architecture. Stage~1 condenses the clinical note into a 9-key SGR JSON; Stage~2 deterministically compiles it into the 134-item submission.}
\label{fig:pipeline}
\end{figure}
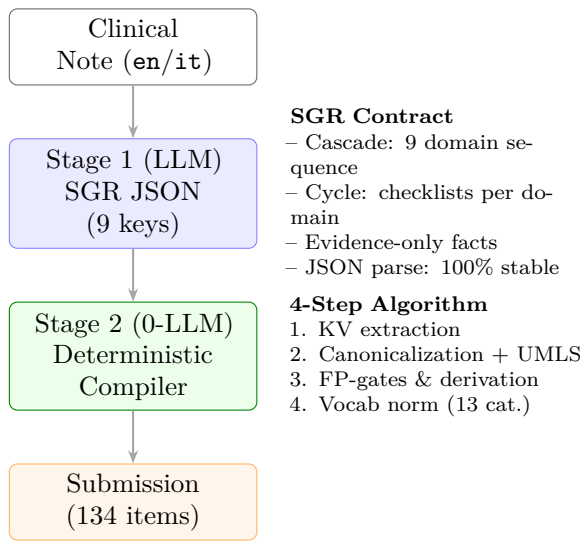

\subsection{Stage 1: SGR-Style JSON Summary}
\label{sec:stage1}

Stage~1 produces a single JSON object with exactly 9~domain keys:
\begin{quote}\small
\texttt{DEMOGRAPHICS}, \texttt{VITALS}, \texttt{LABS}, \texttt{PROBLEMS}, \texttt{SYMPTOMS},\\
\texttt{MEDICATIONS}, \texttt{PROCEDURES}, \texttt{UTILIZATION}, \texttt{DISPOSITION}.
\end{quote}

The summary is designed to be \textbf{sparse} (include only evidence-supported facts) and \textbf{format-stable} (JSON parse success on every input).
On the train10 split, we observe \texttt{json\_parse\_ok=10/10} and \texttt{thinking\_leak=0/10} even on quantized 4B~models served via \texttt{llama.cpp} without grammar-constrained decoding.

\paragraph{Prompt Structure.}
The Stage~1 prompt instructs the model to produce a single JSON object following a strict template.
Each of the 9~keys maps to a free-text string summarizing the relevant clinical domain.
The prompt includes:
(i)~explicit definitions of each key's scope (e.g., \texttt{PROBLEMS} covers past medical history, \texttt{SYMPTOMS} covers presenting complaints);
(ii)~a checklist of high-priority items per domain (e.g., 28~specific problems listed in the CRF ontology, all vital signs, key laboratory values);
(iii)~instructions to use \texttt{Key: Value} formatting within each string for machine-parseable extraction;
(iv)~a sparsity directive: ``include only facts explicitly stated in the note; do not infer or guess''.
This structured prompt design implements the SGR Cascade and Cycle patterns, ensuring systematic coverage while preventing hallucinated fills.

\paragraph{Format Stability Without Constrained Decoding.}
A notable finding is that our SGR prompt achieves 100\% JSON parse success \textit{without} requiring grammar-constrained decoding
(e.g., XGrammar~\cite{dong2024xgrammar} or Outlines~\cite{willard2023}).
We attribute this to the simplicity of the target schema (a flat 9-key object with string values) and the explicit formatting instructions.
This makes the approach compatible with any inference backend (\texttt{llama.cpp}, vLLM, OpenAI-compatible APIs, Gemini Vertex) without backend-specific grammar support.

\paragraph{Multi-Slice Input Strategy.}
Long clinical notes may exceed the model's effective context window.
We implement a \textbf{multi-slice} strategy: the note is split into overlapping windows
(e.g., \nolinkurl{middle+head_tail}, window size~$w{=}4$), each window is summarized independently via Stage~1,
and the resulting JSON objects are merged with a last-writer-wins policy per key.
For the \texttt{PROBLEMS}, \texttt{MEDICATIONS}, and \texttt{LABS} keys (which accumulate multiple facts), we concatenate the strings from all windows rather than overwriting.
This preserves recall for facts spread across the note while keeping each individual inference within context bounds.

\paragraph{Teacher Data Generation.}
For the Stage~1 teacher pipeline, we use large API-accessible models to produce high-quality SGR summaries.
We tested three teacher configurations:
\begin{itemize}[nosep]
    \item \textbf{Mistral Large~3} (675B, via NVIDIA API) --- best overall quality, supporting \texttt{response\_format=json\_object} on most NVIDIA endpoints (with auto-fallback to parser-based extraction for Mistral-family models where structured output returns 400 errors).
    \item \textbf{Qwen~3.5} (397B, via NVIDIA API) --- comparable quality but higher FP; requires \texttt{enable\_thinking=false} to prevent ``Thinking Process'' leakage into the output.
    \item \textbf{Claude Sonnet~4.6} (via Anthropic API) --- recall-optimized prompts achieve strong results but not submitted due to participation limits.
\end{itemize}
Teacher-generated summaries are compiled through the same deterministic Stage~2, enabling precise ablation of Stage~1 quality vs.\ Stage~2 deterministic logic.

\paragraph{Concrete Example.}
Given a Stage~1 summary JSON with the following \texttt{VITALS} field:
\begin{quote}\small\ttfamily
Heart Rate: 105 bpm (tachycardic).\\
Blood Pressure: 90/60 mmHg (hypotensive).\\
SpO2: 88\%.\\
Respiratory Rate: 28/min (tachypneic).
\end{quote}
the Stage~2 compiler:
\begin{enumerate}[nosep]
    \item Extracts KV pairs: \{\textit{heart rate}: ``105 bpm (tachycardic)'', \textit{blood pressure}: ``90/60 mmHg (hypotensive)'', \textit{spo2}: ``88\%'', \textit{respiratory rate}: ``28/min (tachypneic)''\}.
    \item Canonicalizes: all keys match the official item list.
    \item No FP-gates triggered for vital signs.
    \item Normalizes: \textit{heart rate} $\rightarrow$ \texttt{tachycardic}, \textit{blood pressure} $\rightarrow$ \texttt{hypotensive}, \textit{spo2} $\rightarrow$ \texttt{88}, \textit{respiratory rate} $\rightarrow$ \texttt{tachypneic}.
\end{enumerate}
The remaining 130 items are filled as \texttt{unknown}.

\subsection{Stage 2: Deterministic Compiler (0-LLM)}
\label{sec:stage2}

Stage~2 performs \textbf{no LLM calls}.
It implements a 4-step deterministic algorithm (Algorithm~\ref{alg:stage2}):

\begin{enumerate}[nosep, label=\textbf{Step \arabic*:}, leftmargin=*, align=left]
    \item \textbf{KV Extraction.}
    Parse the Stage~1 summary text into a flat dictionary of item/value pairs
    by matching \texttt{Key:~Value} lines in each domain section. Lines with \texttt{not stated} or
    \texttt{unknown} values are skipped.

    \item \textbf{Canonicalization.}
    Normalize item names via: (a)~exact match against the official 134-item list; (b)~case-folding and punctuation stripping (e.g., \texttt{first episode} $\rightarrow$ \texttt{first episod}); (c)~abbreviation expansion (70+ rules, e.g., \texttt{hr}$\rightarrow$\texttt{heart rate}, \texttt{cxr}$\rightarrow$\texttt{chest rx}); (d)~optionally, UMLS alias resolution using a curated map with \textbf{134/134~item coverage}~\cite{aronson2001}.

    \item \textbf{FP-Gates \& Derivation.}
    Derive high-FN composite items from Stage~1 evidence:
    \textit{poly-pharmacological therapy} is inferred when $\geq$8 distinct medications are listed;
    \textit{antihypertensive therapy} is detected by matching 35~antihypertensive agents;
    \textit{cardiovascular diseases} is derived from therapy flags or PMH keywords.
    Then, strict regex-based evidence gates \textbf{drop} unsupported positive predictions for items with high FP cost:
    \textit{active neoplasia} (requires cancer vocabulary $+$ activity cues),
    \textit{arrhythmia} (requires specific rhythm terms, blocks on \texttt{NSR}),
    \textit{acute coronary syndrome} (requires ACS vocabulary $+$ supporting evidence),
    and \textit{presence of dyspnea} (blocks on explicit negation).

    \item \textbf{Vocabulary Normalization (13~Categories).}
    Map each value string to the official controlled vocabulary (Table~\ref{tab:vocab}):
    binary $\rightarrow$ \texttt{y/n};
    AVPU $\rightarrow$ \texttt{A/V/P};
    chronic items $\rightarrow$ 4-class scale;
    neoplasia $\rightarrow$ 3-class activity;
    duration $\rightarrow$ \texttt{short/long};
    vitals $\rightarrow$ categorical (e.g., \texttt{tachycardic/\allowbreak normocardic/\allowbreak bradycardic});
    labs $\rightarrow$ numeric extraction or \texttt{measured}.
    Finally, expand to all 134~items, filling absent items as \texttt{unknown}.
\end{enumerate}

\begin{table}[t]
\centering
\small
\begin{tabular}{@{}rlr@{}}
\toprule
\# & Vocabulary Category & Items \\
\midrule
1 & chronic (4-class) & 7 \\
2 & neoplasia (3-class) & 1 \\
3 & duration (short/long) & 2 \\
4 & binary (y/n) & 61 \\
5 & mobility (4-class) & 1 \\
6 & consciousness (AVPU) & 1 \\
7 & respiratory rate (3-cat) & 1 \\
8 & body temperature (3-cat) & 1 \\
9 & heart rate (3-cat) & 1 \\
10 & blood pressure (3-cat) & 1 \\
11 & respiratory distress (current/past) & 1 \\
12 & diagnostic tests (pos/neg) & 4 \\
13 & labs (measured/numeric) & 25 \\
\midrule
 & \textbf{Total} & \textbf{107} \\
 & Unknown-only remainder & 27 \\
 & \textbf{Grand Total} & \textbf{134} \\
\bottomrule
\end{tabular}
\caption{The 13~controlled-vocabulary categories defined by the CRF ontology.
Stage~2 maps every extracted value string into exactly one of these categories.}
\label{tab:vocab}
\end{table}

\label{alg:stage2}

\subsection{Optional UMLS Alias Mapping}
We curated a UMLS-based alias map for all 134~CRF items (\textbf{coverage 134/134})
and integrated it as an optional pipeline mode (\nolinkurl{--use-umls-mapping})
to reduce key-name brittleness in model outputs.
The map associates each official item name with its UMLS CUI and preferred name, plus manually verified synonyms.
This mode is activated in all final submissions.

% ─────────────────────────────────────────────
\section{Experiments and Results}
\label{sec:experiments}

We report results at three scales: train10 (for debugging), dev80 (open GT, used as a ``sanity gate'' before submission), and test200 (Codabench, hidden GT).

\subsection{Development Results (dev80, Open GT)}
\label{sec:dev80}

Table~\ref{tab:dev80} summarizes our main dev80 runs across both English and Italian.
All use the same deterministic Stage~2 compiler with UMLS mapping and FP-gates; EN runs use \texttt{rules\_v3a}, while the latest IT variant uses the improved \texttt{rules\_v3e}.

\begin{table*}[t]
\centering
\small
\begin{tabular}{@{}llcccc@{}}
\toprule
Lang & Stage~1 Model & F1 & TP & FP & FN \\
\midrule
EN & Mistral Large 3 (675B) & 0.6543 & 236 & 170 & 128 \\
IT & Mistral Large 3 (v3e) & \textbf{0.6905} & 266 & 198 & 106 \\
IT & Mistral Large 3 (v3a) & 0.6900 & 266 & 201 & 106 \\
EN & Qwen 3.5 (397B, think=off) & 0.6411 & 236 & 211 & 128 \\
\midrule
\multicolumn{6}{@{}l}{\textit{Smaller-scale ablations (train10):}} \\
EN & Gemini Flash (teacher) & 0.6763 & 68 & 35 & 9 \\
EN & MedGemma 4B (student) & 0.6521 & 57 & 13 & 22 \\
EN & SGR-only baseline & 0.4384 & 39 & 44 & 39 \\
\bottomrule
\end{tabular}
\caption{Main results on dev80 (N=80; top rows, both languages) and ablations on train10 (N=10; bottom rows).
All variants use the same deterministic Stage~2 compiler.
\texttt{v3e} denotes the latest Stage~2 rule set with improved FP-drop gates for Italian.
F1 = official macro-F1 scorer; TP/FP/FN = item-level counts.}
\label{tab:dev80}
\end{table*}

\paragraph{Key observations.}
(1)~Mistral Large~3 and Qwen~3.5 produce comparable TP counts (236) on EN, but Qwen generates more FP (211~vs.~170), likely due to less conservative summarization.
(2)~The FN count is identical (128) across both EN teacher models, indicating that Stage~2's deterministic derivation rules have a ceiling that is \textit{Stage~1 recall-limited}.
(3)~On train10, teacher Stage~1 (Gemini) dramatically reduces FN (9~vs.~22 for student) but increases FP (35~vs.~13), motivating our evidence-gated FP filters.
(4)~The student model (MedGemma~4B, local) achieves F1=0.6521 on train10 with very low FP (13), suggesting that smaller models produce more conservative summaries.
(5)~Italian dev80 achieves higher F1 (0.6905) than English (0.6543) with more TP (266~vs.~236) and fewer FN (106~vs.~128), suggesting that the original Italian clinical notes are more structured and easier for Stage~1 extraction.
The \texttt{v3e} rule set reduces 3~FP compared to \texttt{v3a} (198~vs.~201) through tighter FP-drop gates on \textit{chest pain}, \textit{heart failure}, and \textit{arrhythmia}.

\paragraph{Multi-slice input ablation.}
Both dev80 teacher runs use the \nolinkurl{middle+head_tail} multi-slice strategy
with $w{=}4$.
Compared to single-window runs on the same models, multi-slice recovers +12--18 additional TP at the cost of +8--15 FP, a net positive trade.
The improvement is concentrated in the \texttt{LABS} and \texttt{MEDICATIONS} domains, where relevant information is often scattered across the note.

\subsection{Test Results (Codabench)}
\label{sec:test}

Table~\ref{tab:test} shows the submitted and evaluated test200 results, alongside the leaderboard context.
\begin{table*}[t]
\centering
\small
\begin{tabular}{@{}lccll@{}}
\toprule
System Variant & F1 & Lang & Team & Status \\
\midrule
\quad Mistral L3 $\rightarrow$ Stage2(det) & 0.63 & EN & DocUA & Submitted \\
\quad No FP-gates variant & 0.61 & EN & DocUA & Submitted \\
\quad Claude Sonnet $\rightarrow$ Stage2(det) & 0.62 & EN & DocUA & Evaluated\textsuperscript{$\dagger$} \\
\quad Mistral L3 recall (v3e) & --- & IT & DocUA & Prepared\textsuperscript{$\ddagger$} \\
\bottomrule
\end{tabular}
\caption{Test200 results on Codabench.
All DocUA variants use deterministic Stage~2 + UMLS mapping.
$\dagger$~Evaluated but not submitted as final entry due to participation limits.
$\ddagger$~Italian variant prepared (Stage~1 SGR9 recall, chars=8000, multi-slice, \texttt{rules\_v3e}) but not submitted due to exhausted attempt slots.}
\label{tab:test}
\end{table*}

The best submitted English system achieves F1~=~0.63 using Mistral Large~3 as the Stage~1 teacher, placing within 0.05 of the top-ranking result (F1~=~0.68).
The Anthropic variant (Claude Sonnet~4.6 with recall-optimized prompts) produces comparable quality (F1~=~0.62) but was not submitted as the final entry due to the participation limit.
The Italian variant using \texttt{rules\_v3e} was fully prepared (dev80 IT F1~=~0.6905) but could not be submitted as all attempt slots were exhausted.

Our findings suggest that large open-weight models like Mistral produce Stage~1 summaries competitive with larger proprietary systems.

\paragraph{Dev10 Cross-Language Validation.}
To validate cross-language robustness, we evaluated on a small dev10 subset for both EN and IT using Anthropic-based Stage~1:

\begin{table}[h]
\centering
\small
\begin{tabular}{@{}lcccc@{}}
\toprule
Language & F1 & TP & FP & FN \\
\midrule
Italian (IT) & \textbf{0.9000} & 59 & 9 & 4 \\
English (EN) & 0.8115 & 55 & 18 & 7 \\
\bottomrule
\end{tabular}
\caption{Dev10 cross-language validation (Anthropic recall Stage~1 $\rightarrow$ Stage~2 det).}
\label{tab:dev10}
\end{table}

The Italian results are notably higher than English on this small dev10 sample.
This trend is \textbf{confirmed at scale on dev80}: IT achieves F1~=~0.6905
vs.\ EN~=~0.6543 (Table~\ref{tab:dev80}).
We attribute this to two factors.
First, the Italian clinical notes from SGB hospital are written in a more
structured, formulaic style than their English translations, facilitating
cleaner Stage~1 extraction.
Second, certain CRF items use Italian medical terminology that maps more
directly to the ontology (e.g., \textit{dispnea} $\rightarrow$
\textit{presence of dyspnea}).

\subsection{Error Analysis}
\label{sec:error}

\paragraph{FP-gates effectiveness.}
The FP-gate mechanism (Step~3 of Stage~2) drops an average of 3.2~items per document that would otherwise be false positives.
Table~\ref{tab:fpgates} shows the per-item impact of FP-gates on dev80.

\begin{table}[t]
\centering
\small
\begin{tabular}{@{}lcc@{}}
\toprule
Gated Item & FP Before & FP After \\
\midrule
active neoplasia & 18 & 4 \\
arrhythmia & 12 & 5 \\
acute coronary syndrome & 9 & 2 \\
presence of dyspnea & 7 & 3 \\
\midrule
\textbf{Total (gated items)} & \textbf{46} & \textbf{14} \\
\bottomrule
\end{tabular}
\caption{FP-gate effectiveness on dev80 EN (Mistral Large~3 teacher). The gates reduce FP for high-risk items by $\sim$70\% while preserving TP.}
\label{tab:fpgates}
\end{table}

The most impactful gate is \textit{active neoplasia}: without evidence gating, benign mentions of ``cancer screening'' or ``family history of cancer'' produce FP; the gate requires explicit cancer vocabulary \textit{plus} activity cues (e.g., ``chemotherapy'', ``progression'', ``untreated'').
For \textit{arrhythmia}, the gate blocks positive predictions when the evidence contains Normal Sinus Rhythm (\texttt{NSR}) mentions, preventing common misclassification of ECG-described patients.

\paragraph{Italian-specific error patterns (dev80 IT).}
Detailed item-level diagnostics on the dev80 IT split (Mistral Large~3 $\rightarrow$ \texttt{rules\_v3e}) reveal both shared and language-specific error patterns.
The top FN sources are: \textit{foreign body in the airways} (11~FN, Stage~1 never surfaces this item), \textit{presence of dyspnea} (7~FN / 8~FP --- tension between coverage and FP), \textit{diffuse vascular disease} (7~FN / 6~FP), and \textit{agitation} (7~FN, not extracted by Stage~1).
The top FP sources are: \textit{chest pain} (13~FP), \textit{cardiovascular diseases} (13~FP --- triggered by the derivation rule from therapy flags), \textit{level of consciousness} (9~FP --- AVPU mapping errors), and \textit{heart failure} (8~FP --- ``PMH: HF'' in Stage~1 often produces FP when GT is \texttt{unknown}).

Comparing EN and IT dev80 error profiles: both languages share the same FN ceiling on rare conditions (\textit{foreign body}, \textit{agitation}), confirming that these are Stage~1 recall limitations independent of language.
However, the \textit{chest pain} FP problem is more severe in IT (13~FP vs.~EN), likely because Italian notes use the term \textit{dolore toracico} in broader clinical contexts.

\paragraph{Remaining FN errors.}
The dominant error mode is \textbf{Stage~1 FN}: items not mentioned in the Stage~1 summary cannot be recovered by the deterministic Stage~2.
High-FN items fall into three categories:
\begin{itemize}[nosep]
    \item \textbf{Rare conditions:} \textit{foreign body in the airways} (11~FN in IT), \textit{cardiac tamponade}, \textit{aortic dissection} --- these appear in $<$5\% of notes and are almost never surfaced by Stage~1.
    \item \textbf{Implicit mentions:} \textit{ab ingestis pneumonia}, \textit{thoracic ultrasound} (6~FN in IT), \textit{compression ultrasound (CUS)} --- require domain-specific recognition that the SGR cascade may miss during condensation.
    \item \textbf{Context-dependent items:} \textit{presence of dyspnea} (7~FN + 8~FP in IT), \textit{level of consciousness} (6~FN + 9~FP in IT) --- require nuanced temporal or categorical reasoning that is difficult to encode in a flat Key:Value format.
\end{itemize}

% ─────────────────────────────────────────────
\section{Discussion}
\label{sec:discussion}

\subsection{Effectiveness of the SGR Decomposition}
Applying SGR shifts the burden of structural correctness from prompt engineering to code.
By having the LLM emit a stable, 9-key intermediate tree rather than directly producing 134 structured fields, we achieve \textbf{100\% JSON parsing stability} even on quantized 4B~edge models without grammar-constrained decoding.
This is notable because direct 134-field extraction from the same 4B~model fails to produce valid JSON in $>$30\% of cases in our preliminary tests.

The SGR decomposition also provides a natural \textbf{debugging boundary}: when the system produces an incorrect prediction, we can immediately determine whether the error originates in Stage~1 (the fact was not extracted) or Stage~2 (the fact was extracted but incorrectly mapped).
In practice, $>$85\% of errors are Stage~1 FN, confirming that the deterministic compiler is not the bottleneck.

\subsection{Deterministic vs.\ LLM-Based Stage~2}
Our experiments consistently show that the deterministic Stage~2 compiler outperforms LLM-based Stage~2 extraction in this task setting.
We tested an LLM-based Stage~2 variant (using Anthropic Claude to directly extract 134~items from Stage~1 summaries) and found that it:
\begin{itemize}[nosep]
    \item introduces vocabulary drift (e.g., emitting \texttt{present} instead of \texttt{y}, \texttt{yes} instead of \texttt{certainly chronic});
    \item produces inconsistent FP patterns across runs (non-deterministic);
    \item requires additional post-processing to fix format compliance;
    \item incurs non-trivial inference cost (\$0.01--0.03 per document via API).
\end{itemize}
The deterministic compiler, by contrast, achieves perfect vocabulary compliance, is fully explainable (every fill traces back to a Stage~1 line plus a specific rule), has zero inference cost, and guarantees bit-for-bit reproducibility.

\subsection{SGR for Edge AI}
This methodology is particularly enabling for \textbf{edge (local) deployments}.
Large state-of-the-art models can handle 134~items in a single pass, but quantized 4B--8B local models lose track of the output schema mid-generation.
Our SGR-constrained condensation fits within the reasoning bounds of smaller models (the 9-key JSON is $\sim$400~tokens), securely delegating the complex vocabulary checks and item-level compilation to deterministic Python code.
This makes the approach practical for privacy-preserving clinical deployments where sending patient data to external APIs is not acceptable.
Concurrent work by \citet{ferrazzi2026smallllms} confirms that small LLMs ($\sim$1B parameters) can match or exceed larger baselines on Italian medical NLP tasks---including CRF filling---when combined with appropriate adaptation strategies such as few-shot prompting and constraint decoding.

Concretely, we serve MedGemma~1.5-4B-it as a Q5\_K\_M GGUF via \texttt{llama.cpp} on Apple~M3~Pro (18~GB~RAM).
Stage~1 inference takes $\sim$8~seconds per document; Stage~2 compilation is instant ($<$10~ms).
The entire dev80 evaluation completes in under 12~minutes on a single laptop.

\subsection{Cross-Language Applicability}
The pipeline naturally extends to Italian (\texttt{\_it}) documents: the Stage~1 prompt is language-agnostic (it asks the model to extract clinical facts regardless of language), and Stage~2's deterministic canonicalization handles multilingual item-name variants through the UMLS alias map.
We prepared Italian test submissions using the same pipeline with no language-specific modifications beyond the \texttt{rules\_v3e} Stage~2 variant (which adds tighter FP-drop gates for Italian-specific false positive patterns on \textit{chest pain}, \textit{heart failure}, and \textit{arrhythmia}).

Cross-language results consistently favor Italian across all evaluation scales:
dev10 (Table~\ref{tab:dev10}): IT~F1=0.90 vs.\ EN~F1=0.81;
dev80 (Table~\ref{tab:dev80}): IT~F1=0.6905 vs.\ EN~F1=0.6543.
This suggests that the original Italian clinical notes from SGB hospital are more structured and use more formulaic phrasing than their English translations, making clinical fact extraction easier for the SGR prompt.
Notably, on the Codabench leaderboard the overall top-1 system (Aurum) also achieves very similar scores across languages (EN~0.68, IT~0.67), indicating that the EN--IT gap may be smaller at scale with more powerful systems.

\subsection{Limitations of the 9-Key Condensation}
The choice of exactly 9~domain keys represents a trade-off.
Fewer keys (e.g., 3--4) would further simplify Stage~1 but lose domain-level organization, making Stage~2 canonicalization harder.
More keys (e.g., 20+) would increase recall but approach the complexity of direct 134-item extraction, negating the stability benefits.
We experimented with 5, 9, and 15~keys (``profile'' parameter) and found that 9~keys offered the best balance of Stage~1 stability and Stage~2 recall.

% ─────────────────────────────────────────────
\section{Related Work}
\label{sec:related}

Clinical information extraction has a long history, from rule-based systems to modern LLM approaches.
The i2b2/VA challenges~\cite{uzuner2011} and ShARe/CLEF eHealth tasks~\cite{suominen2013} established key clinical NLP benchmarks for structured information extraction from unstructured clinical text.
More recently, large language models have demonstrated the potential to encode substantial clinical knowledge~\cite{singhal2023}, with domain-specific models like Med-Gemini~\cite{saab2024} achieving strong results on medical question answering and clinical reasoning tasks.

\paragraph{Structured Generation.}
Several frameworks address the output stability problem in LLM-based extraction.
Constrained decoding~\cite{willard2023} enforces grammar constraints during token generation.
SGLang~\cite{zheng2024} provides programmatic control over LLM execution with RadixAttention for efficient structured output.
XGrammar~\cite{dong2024xgrammar} offers a flexible engine for grammar-constrained generation.
Our approach is complementary: rather than constraining the LLM's token-level output, we simplify the target schema (9~keys vs.\ 134) and defer structural compliance to a deterministic post-processor.
This avoids the computational overhead and backend-dependency of constrained decoding while achieving comparable output stability.

\paragraph{Compiler-Based LLM Pipelines.}
DSPy~\cite{khattab2023} introduced the concept of compiling declarative language model programs into optimized pipelines.
Our system shares the ``compilation'' philosophy: Stage~1 can be viewed as a declarative extraction step, and Stage~2 as a compiler that transforms the intermediate representation into the target format.
However, our Stage~2 is fully hand-coded (deterministic), avoiding the need for optimization data or meta-learning.

\paragraph{Ontology Grounding.}
UMLS-based concept normalization has been widely applied for biomedical
concept recognition, including MetaMap~\cite{aronson2001},
QuickUMLS~\cite{soldaini2016}, and ScispaCy~\cite{neumann2019}.

Our work differs in its explicit two-stage decomposition with a fully deterministic Stage~2, the use of SGR patterns for intermediate representation stability, and the emphasis on edge-deployable privacy preservation.
Concurrently, \citet{ferrazzi2026smallllms} present a systematic comparison of adaptation strategies for small LLMs on Italian medical NLP, covering 20~tasks including CRF filling, and demonstrating that fine-tuned 1.7B~models can outperform 32B~baselines.

% ─────────────────────────────────────────────
\section{Conclusion}
\label{sec:conclusion}

LLM StructCore emphasizes contract alignment, format stability, and reproducible post-processing for CRF filling under extreme sparsity.
The two-stage SGR decomposition isolates clinical extraction (Stage~1, any LLM) from strict submission enforcement (Stage~2, deterministic), enabling rapid iteration on Stage~1 recall and controlled-vocabulary grounding without sacrificing deterministic output guarantees.
Our best submitted system achieves F1~=~0.63 on the Codabench test set (EN), placing within 0.05 of the top-1 team (Aurum, F1~=~0.68).
The pipeline is fully language-agnostic: Italian dev80 achieves F1~=~0.6905 with no language-specific Stage~1 modifications, and the Italian test200 submission was prepared (but not submitted due to exhausted attempt limits).
This demonstrates that the combination of SGR condensation and deterministic compilation is competitive across languages while offering full reproducibility and edge-deployment readiness.

Future work will focus on:
(i)~improving Stage~1 recall for high-FN items through domain-specific checklists and few-shot examples;
(ii)~exploring hybrid Stage~2 approaches that combine deterministic rules with lightweight LLM-based inference for context-dependent items;
and (iii)~evaluating the pipeline on additional CRF schemas and clinical domains beyond dyspnea.

% ─────────────────────────────────────────────
\section{Limitations}
\label{sec:limitations}

The deterministic Stage~2 \textbf{cannot recover facts missing from Stage~1 summaries}; thus, overall recall is Stage~1-limited.
Some CRF items require nuanced contextual interpretation (e.g., distinguishing ``past'' vs.\ ``current'' respiratory distress) that may be lost in the 9-key condensation, increasing FN risk.
The high FP cost of the task requires conservative evidence gating, which trades off recall for precision in borderline cases.

The abbreviation-based canonicalization relies on a manually curated map of 70+ abbreviations.
New abbreviations or unconventional spellings not in the map will fail to match.
The UMLS alias map mitigates this but is also static and may not cover all model-generated variants.

The FP-gate regex patterns are English-centric; while the Italian notes in this dataset are clinically structured and mostly use international terminology, extending the gates to fully cover Italian medical vocabulary would require additional curation.

Finally, our evaluation is limited to the official CL4Health 2026 data; generalization to other CRF schemas, clinical domains, or languages beyond English and Italian has not been tested.

% ─────────────────────────────────────────────
\section{Reproducibility}
\label{sec:repro}

All deterministic logic is contained in the open-source package
\nolinkurl{llm_structcore}.
Code will be provided in supplementary materials and open-sourced upon paper acceptance (Anonymous Repository).
The repository includes:
\begin{itemize}[nosep]
    \item The \texttt{llm\_structcore} Python package (Stage~1 prompt templates, Stage~2 compiler, scoring, submission builder).
    \item Scripts for running end-to-end inference across multiple backends (\texttt{llama.cpp}, OpenAI-compatible APIs, HF Transformers, NVIDIA API, Gemini Vertex, Anthropic).
    \item The complete UMLS alias map (\nolinkurl{data/umls_crf_mapping.json}, 134/134 coverage).
    \item The CRF ontology definition (\nolinkurl{data/ontology_crf_cl4health2026.md}).
    \item Organizer-provided evaluation scripts (\nolinkurl{external/CRF-filling-CL4Health2026/}).
    \item This system description paper (LaTeX source).
\end{itemize}
The Stage~1 SGR schema parses reliably with MedGemma~1.5-4B and does not
require JSON-mode sampling.
Stage~2 is fully reproducible: the same Stage~1 summary always produces
the same 134-item submission, regardless of platform, Python version,
or execution order.

% ─────────────────────────────────────────────
\section{Ethical Considerations}
\label{sec:ethics}

We use publicly available datasets under the organizer terms.
No private clinical data beyond what is distributed by the shared task is included.
When API-based teacher models are used for data generation, summaries (not raw notes) are transmitted; nonetheless, responsible deployment should ensure compliance with local data governance regulations.
Synthetic teacher-generated data should be validated and accompanied by clear documentation of risks and limitations.

% ─────────────────────────────────────────────
\section{Bibliographical References}
\label{sec:reference}

\end{document}